\documentclass[conference]{IEEEtran}
\IEEEoverridecommandlockouts
\usepackage{cite}
\usepackage{amsmath,amssymb,amsfonts}
\usepackage{graphicx}
\usepackage{textcomp}
\usepackage{xcolor}
\usepackage{algorithm}
\usepackage[noend]{algpseudocode}
\algrenewcommand{\algorithmiccomment}[1]{\hskip1em$\#$ #1}
\usepackage[
  bookmarks=false,
  pdfpagelabels=false,
  hyperfootnotes=false,
  hyperindex=false,
  pageanchor=false,
  colorlinks=false
]{hyperref}
\usepackage{url}

\def\BibTeX{{\rm B\kern-.05em{\sc i\kern-.025em b}\kern-.08em
    T\kern-.1667em\lower.7ex\hbox{E}\kern-.125emX}}

\begin{document}

\title{Solution and Fitness Evolution (SAFE):\\A Study of Multiobjective Problems\\
\thanks{\textcopyright Copyright 2019 IEEE.
Proceedings of 2019 IEEE Congress on Evolutionary Computation (CEC).
This work was supported by National Institutes of Health (USA) grants AI116794, LM010098, and LM012601.}
}

\author{
\IEEEauthorblockN{Moshe Sipper}
\IEEEauthorblockA{\textit{Institute for Biomedical Informatics} \\
\textit{University of Pennsylvania}\\
Philadelphia, PA 19104-6021, USA \\
and \textit{Department of Computer Science} \\
\textit{Ben-Gurion University} \\
Beer Sheva 84105, Israel \\
sipper@upenn.edu}

\and
\IEEEauthorblockN{Jason H. Moore}
\IEEEauthorblockA{\textit{Institute for Biomedical Informatics} \\
\textit{University of Pennsylvania}\\
Philadelphia, PA 19104-6021, USA \\
jhmoore@upenn.edu}

\and
\IEEEauthorblockN{Ryan J. Urbanowicz}
\IEEEauthorblockA{\textit{Institute for Biomedical Informatics} \\
\textit{University of Pennsylvania}\\
Philadelphia, PA 19104-6021, USA \\
ryanurb@upenn.edu}

}

\maketitle

\begin{abstract}
We have recently presented SAFE---Solution And Fitness Evolution---a commensalistic coevolutionary algorithm that maintains two coevolving populations: a population of candidate solutions and a population of candidate objective functions. We showed that SAFE was successful at evolving solutions within a robotic maze domain. Herein we present an investigation of SAFE's adaptation and application to multiobjective problems, wherein candidate objective functions explore different weightings of each objective. Though preliminary, the results suggest that SAFE, and the concept of coevolving solutions and objective functions, can identify a similar set of optimal multiobjective solutions without explicitly employing a Pareto front for fitness calculation and parent selection. These findings support our hypothesis that the SAFE algorithm concept can not only solve complex problems, but can adapt to the challenge of problems with multiple objectives.  
\end{abstract}


\section{Introduction}
\label{sec:intro}

We have recently highlighted a fundamental problem recognized to confound algorithmic optimization, namely, \textit{conflating} the objective with the objective function \cite{Sipper2018}. Even when the former is well defined, the latter may not be obvious. We presented an approach to automate the means by which a good objective function may be discovered, through the introduction of SAFE---\textbf{S}olution \textbf{A}nd \textbf{F}itness \textbf{E}volution (\textbf{SAFE})---a \textit{commensalistic} coevolutionary algorithm that maintains two coevolving populations: a population of candidate solutions and a population of candidate objective functions \cite{Sipper2019}. We showed that SAFE successfully evolved solutions within a robotic maze domain, simultaneously with the objective functions needed to measure solution quality during evolution.

In this paper we perform a preliminary investigation into SAFE's efficacy in solving multiobjective problems, targeting the well-known ZDT benchmark suite. 
With multiobjective problems the evolutionary algorithm is tasked with optimizing more than one objective simultaneously. The objective function---which, as noted above, should not be conflated with the objective(s)---thus needs to perform a ``juggling act'' that balances the (multiple) objectives in an efficacious manner, producing viable solutions. SAFE, with its inherent ability to devise its own objective functions on the fly (as we shall see below), is thus a good option to try out in the domain of multiobjective optimization.

Ultimately, our goal is to employ this SAFE algorithm concept as a machine learning (ML) algorithm for the modeling and discovery of complex patterns of associations, which manifest themselves as problems with a deceptive fitness landscape. ML has been defined as computer programs (i.e., algorithms) where the performance (P) on some task (T) is improved with experience (E) \cite{mitchell1997machine}. However, this makes the assumption that the P metric is both fixed and it has been correctly selected a priori to represent T optimality. This assumption is often violated in real-world problems leading to bias and overfitting. 

In the next section we provide a pr\'ecis of multiobjective optimization. 
Section~\ref{sec:coevolution} discusses coevolutionary algorithms. 
Section~\ref{sec:safe} presents the SAFE algorithm and describes how it can be applied to the current domain of interest. 
Note that, as opposed to many other multiobjective optimizers, SAFE does not rely on measures of the Pareto front (i.e., a Pareto front is not employed to calculate solution fitness, or as a standard for selecting parent solutions to generate offspring solutions).
Experiments and results are provided in Section~\ref{sec:results}, followed by concluding remarks in Section~\ref{sec:conc}.

\section{Multiobjective Optimization} 
\label{sec:multiobj}

A multiobjective optimization problem involves two or more objectives all of which need to be optimized. 
For example, think of buying a car, where the objectives are to optimize both cost and fuel efficiency. 
Applications of multiobjective optimization abound in numerous domains \cite{zhou2011multiobjective}.

Consider the 2-objective ZDT problems, defined in Table~\ref{tab:problems} \cite{zitzler2000comparison,huband2006review}.
The two objectives are to minimize both $f_1(\textbf{x})$ and $f_2(\textbf{x})$. This classic benchmark suite epitomizes the basic setup of multiobjective optimization. The utility of this suite is that the ground-truth optimal Pareto front can be computed and used to determine and compare multiobjective algorithm performance. 

\begin{table*}
\centering
\caption{Two-objective benchmark ZDT problems. 
The two objectives are to minimize both $f_1(\textbf{x})$ and $f_2(\textbf{x})$.
The dimensionality of the problem is $k=30$, i.e.,
solution vector 
$\textbf{x} = x_1,\ldots,x_{30}$}\label{tab:problems}
\bgroup
\def\arraystretch{1.7}
\begin{tabular}{|r|l|c|}
\hline
Name  & Problem & Parameter domain \\ \hline
ZDT1  & $ f_1(\textbf{x}) = x_1 $ & $[0,1]$ \\
      & $ g(\textbf{x}) = 1 + 9/(k-1)\sum_{i=2}^{k} x_i $ & \\
      & $ f_2(\textbf{x}) = 1-\sqrt{f_1/g} $ & \\ \hline 
ZDT2  & as ZDT1, except $ f_2(\textbf{x}) = 1-(f_1/g)^2 $ & $[0,1]$\\ \hline 
ZDT3  & as ZDT1, except $ f_2(\textbf{x}) = 1-\sqrt{f_1/g} - (f_1/g)\sin(10 \pi f_1) $ & $[0,1]$ \\ \hline 
ZDT4  & as ZDT1, except $ g(\textbf{x}) = 1 + 10(k-1) + \sum_{i=2}^{k} (x_i^2 - 10 \cos(4\pi x_i) ) $ & $x_1 \in [0,1]$, $x_2,\ldots,x_n \in [-5,5]$ \\ \hline 
\end{tabular}
\egroup
\end{table*}

With a multiobjective optimization problem there is usually no single-best solution, but rather the goal is to identify a set of `non-dominated' solutions that represent optimal tradeoffs between multiple objectives. This set of solutions is also known as the \textit{Pareto optimal set}. Usually, a representative subset will suffice.

Reviewing the state of the art, Zhou et al. \cite{zhou2011multiobjective} noted several types of multiobjective algorithms: 1) those based on decomposition, where the problem is decomposed into a number of scalar objective optimization problems; 2) preference-based algorithms, where the user provides information about preferred solutions; 3) indicator-based algorithms, where the quality of an approximated Pareto front is measured by a scalar indicator such as generational distance or hypervolume, which is used to guide the search; 4) hybrid algorithms, which hybridize different search methods; 5) memetic multiobjective algorithms, which incorporate local search; and 6) coevolutionary methods, which is the category our method fits in.

\section{Coevolution}
\label{sec:coevolution}

\begin{figure*}
\centering
\begin{tabular}{ccc}
\includegraphics[height=0.21\textwidth]{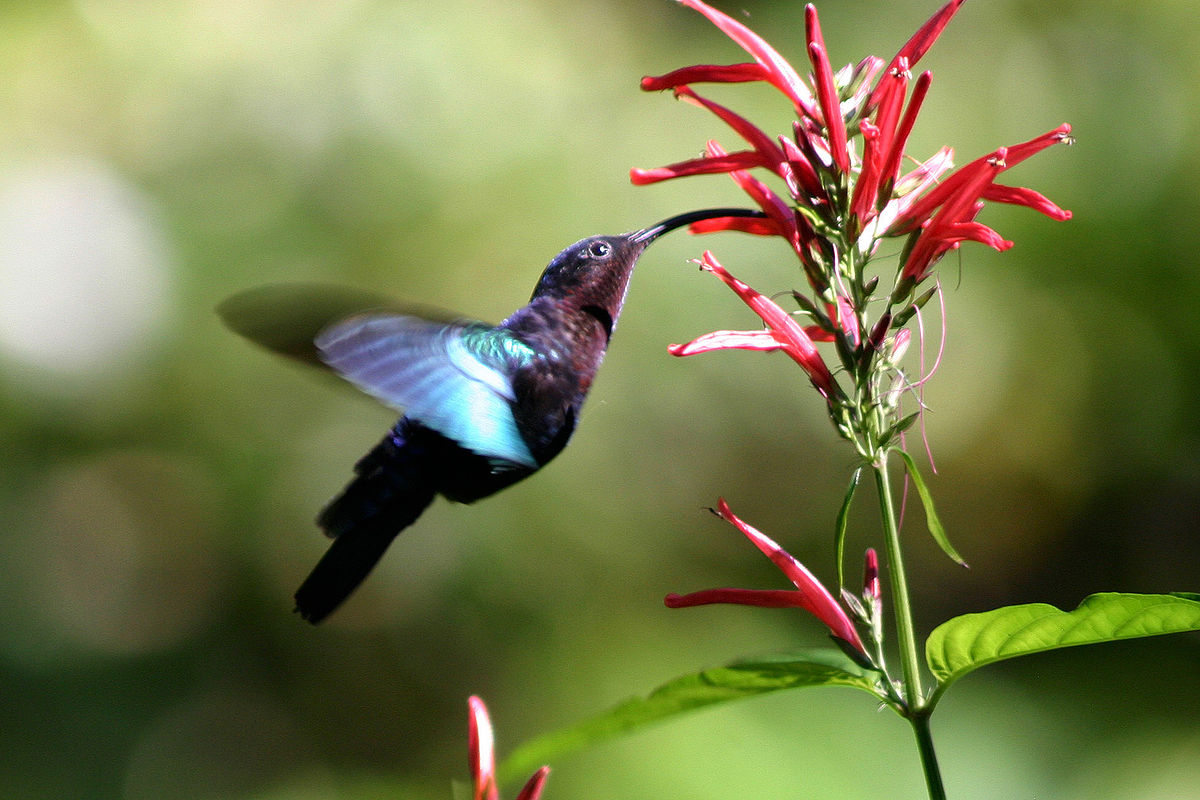} &
\includegraphics[height=0.21\textwidth]{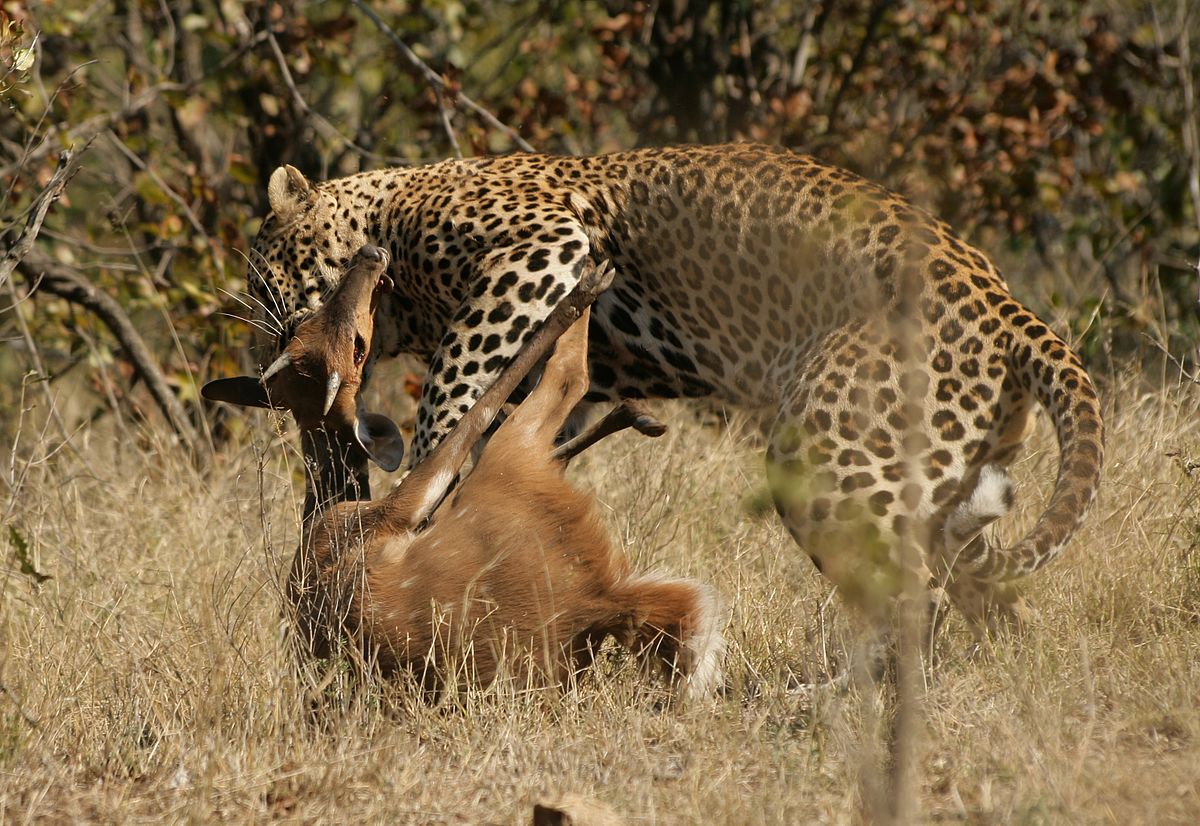} &
\includegraphics[height=0.21\textwidth]{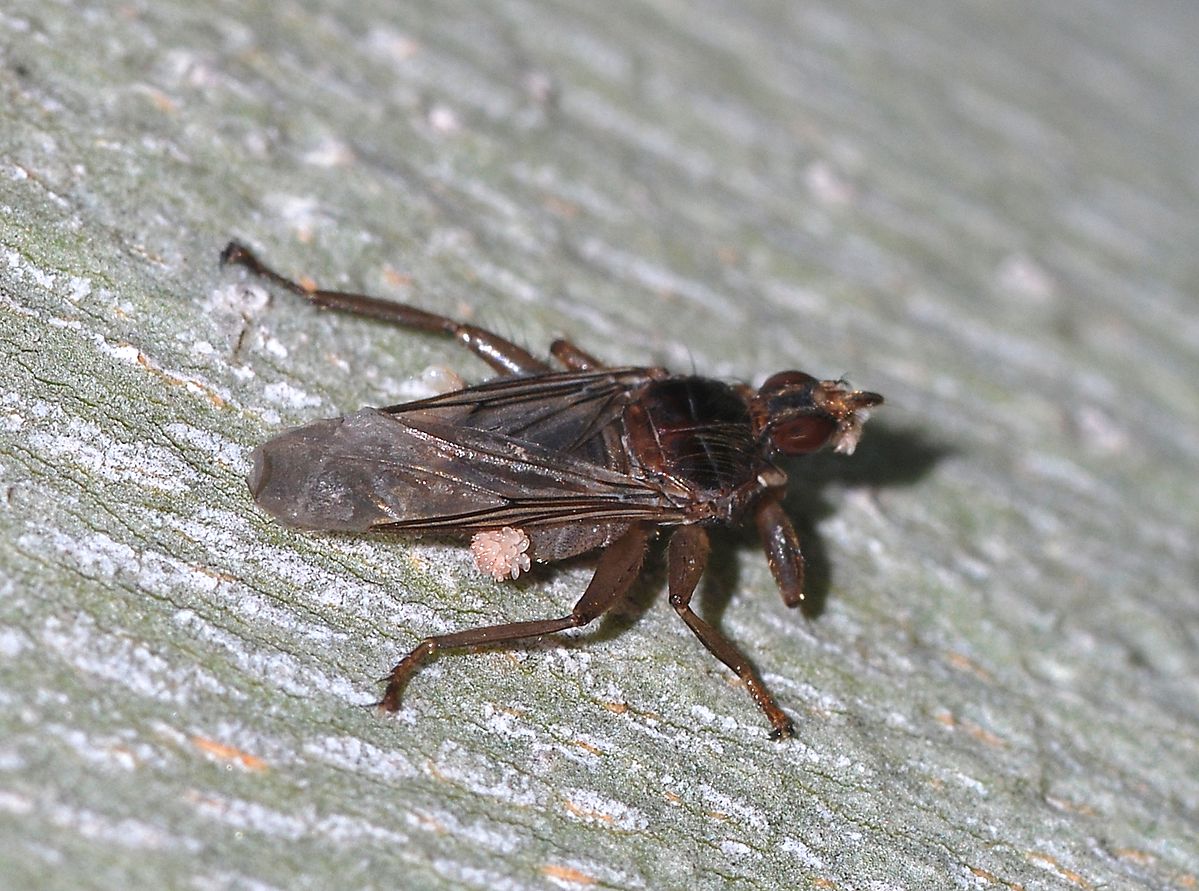} \\
(a) & (b) & (c) \\
\end{tabular}
\caption{Coevolution: 
(a) cooperative: Purple-throated carib feeding from and pollinating a flower (credit: Charles J Sharp,
\url{https://commons.wikimedia.org/wiki/File:Purple-throated_carib_hummingbird_feeding.jpg});
(b) competitive: predator and prey---a leopard killing a bushbuck
(credit: NJR ZA, \url{https://commons.wikimedia.org/wiki/File:Leopard_kill_-_KNP_-_001.jpg});
(c) commensalistic: Phoretic mites attach themselves to a fly for transport
(credit: Alvesgaspar, \url{https://en.wikipedia.org/wiki/File:Fly_June_2008-2.jpg}).
} \label{fig:coevolution}
\end{figure*}

Coevolution refers to the simultaneous evolution of two or more species with coupled fitness \cite{Pena:2001}. Strongly related to the concept of symbiosis, coevolution can be mutualistic, parasitic, or commensalistic \cite{wiki:Symbiosis}:
1) In mutualism, different species exist in a relationship in which each individual (fitness) benefits from the activity of the other; 2) in parasitism, an organism of one species competes with an organism of a different species; and 3) in commensalism, members of one species gain benefits while those of the other species neither benefit nor are harmed.
The idea of coevolution originates (at least) with Darwin---who spoke of ``coadaptations of organic beings to each other'' \cite{Darwin:1859}  (Figure~\ref{fig:coevolution}).

A cooperative (mutualistic) coevolutionary algorithm involves a number of independently evolving species, which come together to obtain problem solutions. The fitness of an individual depends on its ability to collaborate with individuals from other species \cite{zaritsky2004,Pena:2001,Potter:2000,Dick:2014}. 

In a competitive (parasitic) coevolutionary algorithm the fitness of an individual is based on direct competition with individuals of other species, which in turn evolve separately in their own populations. Increased fitness of one of the species implies a reduction in the fitness of the other species \cite{Hillis:1990}. 

To the best of our knowledge, SAFE is the first coevolutionary algorithm to employ a form of commensalism \cite{Sipper2019}.

\section{SAFE} 
\label{sec:safe}

\begin{figure*}
\centering
\includegraphics[width=0.97\textwidth]{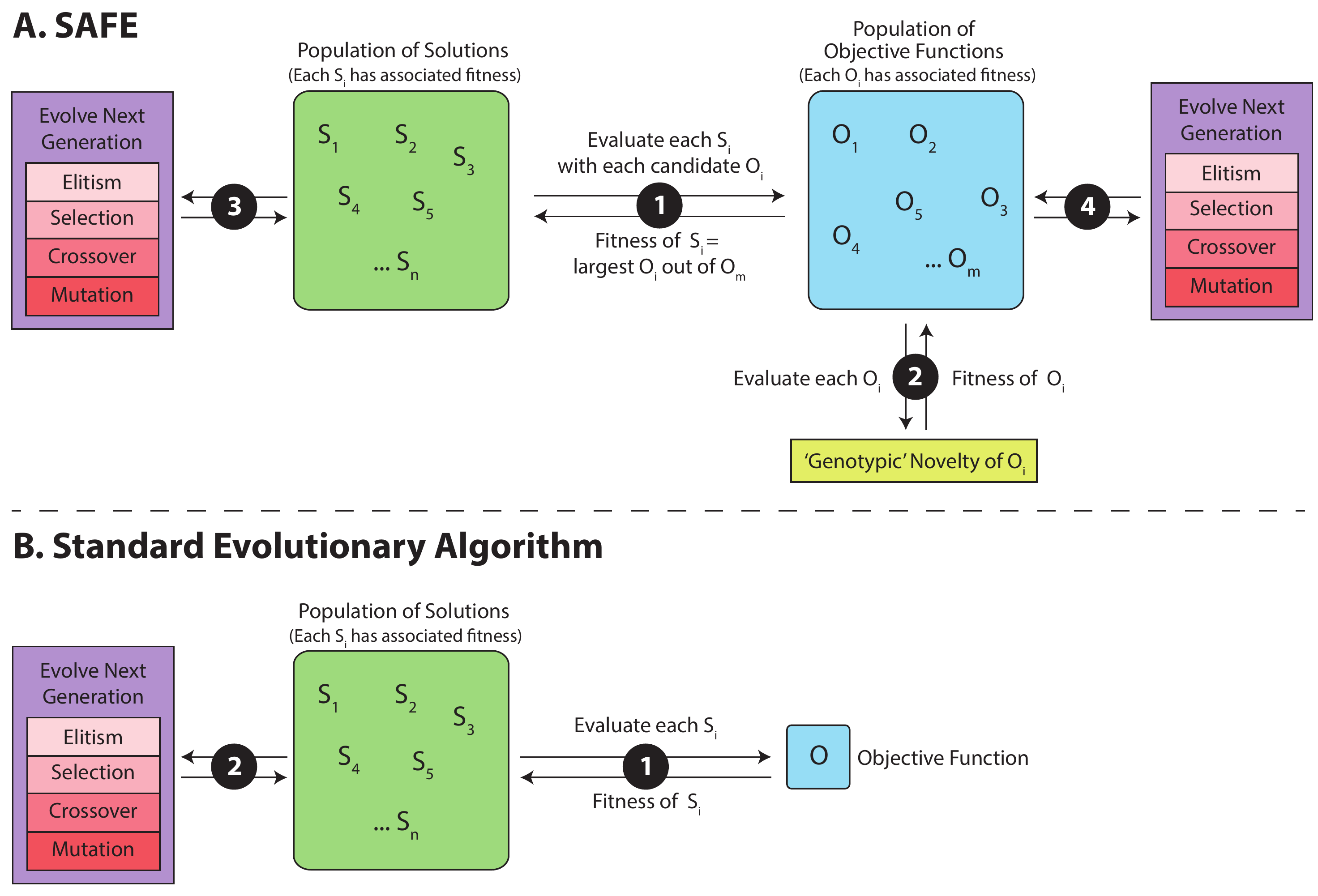} 
\caption{A single generation of SAFE vs. a single generation of a standard evolutionary algorithm. The numbered circles identify sequential steps in the respective algorithms. 
The objective function can comprise a single or multiple objectives.
Recall, also, that we advocate not conflating objective and objective function.} \label{fig:safe}
\end{figure*}

SAFE is a coevolutionary algorithm that maintains two coevolving populations: a population of candidate solutions and a population of candidate objective functions (Figure~\ref{fig:safe}).
The evolution of each population is identical to a standard, single-population evolutionary algorithm---except where fitness computation is concerned.
Below we describe the various components of the system, which includes: (1) population composition, (2) initialization, (3) selection, (4) elitism, (5) crossover, (6) mutation, and (7) fitness computation.

\textbf{Populations.}
SAFE maintains two coevolving populations. An individual in the \emph{solutions population} is a list of 30 real values, each in the range given in Table~\ref{tab:problems} and defined by the respective ZDT problems. .

An individual in the \emph{objective-functions population} is a list of 2 real values $[a,b]$, each in the range $[0,1]$, whose usage is described under \emph{fitness} below. Population sizes and other parameters are given in Table~\ref{tab:params} (Sipper et al. \cite{Sipper2018par} provide interesting insights into parameters in evolutionary computation).

\begin{table}
\centering
\caption{Evolutionary parameters}\label{tab:params}
\bgroup
\def\arraystretch{1.4}\begin{tabular}{|r|l|}
\hline
Description & Value \\ \hline
Number of evolutionary runs & 50  \\
Maximal number of generations & 3000\\
Size of solutions population & 500 \\
Size of objective-functions population & 150 \\
Type of selection & Tournament \\
Tournament size & 5 \\
Type of crossover & single-point \\
Crossover rate & 0.8 \\
Probability of mutation (solutions) & 0.4 \\
Probability of mutation (objective functions) & 0.4 \\
Number of top individuals copied (elitism) & 2 \\ \hline
\end{tabular}
\egroup
\end{table}

\textbf{Initialization.}
For every evolutionary run: both populations are initialized to random (fixed-length) lists, wherein each component value is in the appropriate range.

\textbf{Selection.} 
Tournament selection with tournament size 5, i.e., choose 5 individuals at random from the population and return the individual with the best fitness as the selected one.

\textbf{Elitism.} The 2 individuals with the highest fitness in a generation are copied (``cloned'') into the next generation unchanged.

\textbf{Crossover.} Standard single-point crossover, i.e., select a random crossover point and swap two parent genomes beyond this point to create two offspring. The crossover rate is the probability with which crossover between two selected parents occurs. (Note that for an objective-function individual, which comprises two values, if crossover occurs, it will always be, ipso facto, at the same position.)

\textbf{Mutation.} 
Mutation is done with probability 0.4 (per individual in the population) by selecting a random gene (of the 30 or 2, respectively) and replacing it with a new random value in the appropriate range. 

\textbf{Fitness.}
Fitness computation is where SAFE dynamics come into play. 
In SAFE, each solution individual, $S_i$, $i \in \{1,\ldots,n\}$ is scored by every candidate objective-function individual, $O_j$, in the current population, $j \in \{1,\ldots,m\}$ (Figure~\ref{fig:safe}A). This is in contrast with a standard evolutionary algorithm, where a population of solutions is evolved but the objective function is fixed (Figure~\ref{fig:safe}B).

A candidate SAFE objective-function individual describes a candidate set of weights, balancing the two objectives of the ZDT functions. The best (highest) of all calculated objective function scores (out of the set of all individual objective functions in the objective-function population) is then assigned to the individual solution as its fitness value.

As noted above, an objective-function individual is a pair $[a,b]$; specifically, $a$ determines $f_1$'s weighting and $b$ determines $f_2$'s weighting. $O_j(S_i)$ is the fitness score that objective function $O_j$ assigns to solution $S_i$.
Algorithm~\ref{alg:sol-fit} details fitness computation of solutions. 

\begin{algorithm}
\caption{Compute fitness values of solutions population}\label{alg:sol-fit}
\begin{algorithmic}[1]
\State $n$ $\gets$ size of solutions population
\State $m$ $\gets$ size of objective-functions population
\For{$i \gets 1$ to $n$} 
\State compute $f_1(S_i)$, $f_2(S_i)$
\EndFor

\For{$i \gets 1$ to $n$} 
\For{$j \gets 1$ to $m$} 
\State $a'_j \gets a_j/(a_j+b_j)$
\State $b'_j \gets b_j/(a_j+b_j)$
\State $O_j(S_i) \gets 1/(a'_j \times f_1(S_i) +    
                       b'_j \times f_2(S_i))$ 
\EndFor
\State $\mathit{solutionFitness_i} \gets \max_j O_j$
\EndFor
\end{algorithmic}
\end{algorithm}

As for the objective-functions population, determining the quality of an evolving objective function places us in uncharted waters. Such an individual is not a solution to a problem, but rather the ``guide''---or ``path''---to a solution. As such, it is not clear what comprises a good measure of success. As in \cite{Sipper2019}, we turned to a commensalistic coevolution strategy, where the objective functions' fitness does not depend on the population of solutions. Instead, it relies on \textit{genotypic novelty}.

Novelty search was introduced by \cite{Lehman2008}, the key idea being that instead of rewarding closeness to objective, individual solutions would be considered valuable if their behavior diverged from prior solutions. Effectively, novelty search is identical to a standard evolutionary algorithm, with the fitness function replaced by the novelty metric. We seek novel behaviors rather than the objective, hoping that the former will ultimately lead to the latter.

In our case, the novelty metric is based on the objective-function individual's two-valued genome, $[a,b]$. The distance between two objective functions---$[a_1,b_1]$, $[a_2,b_2]$---is simply the Euclidean distance of their genomes, given as: $\sqrt{(a_1-a_2)^2 + (b_1-b_2)^2}$. Note that this is a genotypic novelty metric, defined by referring to the genome, as opposed to phenotypic, behavioral novelty, used elsewhere \cite{Sipper2019}.

Each generation, every candidate objective function is compared to its cohorts in the current population of objective functions and to an archive (of size 1000) of past individuals whose behaviors were highly novel when they emerged. The novelty score is the average of the distances to the $k$ ($=15$) nearest neighbors, and is used in computing objective-function fitness (Algorithm~\ref{alg:obj-fit}). 

\begin{algorithm}
\caption{Compute fitness values of objective-functions population}\label{alg:obj-fit}
\begin{algorithmic}[1]
\State $m$ $\gets$ size of objective-functions population
\For{$i \gets 1$ to $m$} 
\State compute $\mathit{noveltyScoreObj_i}$
\State $\mathit{objectiveFitness_i} \gets \mathit{noveltyScoreObj_i}$ 
\EndFor
\end{algorithmic}
\end{algorithm}

Every generation each new solution is examined and added to a Pareto front if appropriate, this front being the algorithm's final desired output. Note, however, that while the front is continually updated for tracking and analysis purposes, it is \textit{not} used in any way by SAFE in fitness calculation.

\section{Experiments and Results} 
\label{sec:results}

We performed 50 replicate runs of SAFE for each of the four ZDT problems, ZDT1, ZDT2, ZDT3, ZDT4, recording the evolving Pareto front as evolution progressed. In particular, we were interested in the \textit{inverted generational distance (igd)} of the Pareto fronts evolved.

The $\mathit{igd}$ is one of the leading measures used in recent years to asses the success of a multiobjective optimizer. It measures the distance between the true Pareto front and the  front produced by an algorithm \cite{coello2004study}. The metric is defined as:
\[
\mathit{igd} = \frac{\sum_{v \in \mathit{TF}}d(v,\mathit{PF})}
                    {|\mathit{TF}|}
\]
where $\mathit{TF}$ is the true Pareto front, $\mathit{PF}$ is the front found by the algorithm, and $d(v,\mathit{PF})$ is the minimum Euclidean distance (measured in objective space) between $v$ and the solutions in $\mathit{PF}$; $|\mathit{TF}|$ is the cardinality of set $\mathit{TF}$. The $\mathit{igd}$ measures both the convergence and spread of the obtained Pareto solutions along the true Pareto front \cite{cheng2017novel}.

Table~\ref{tab:results} shows the average inverted generational distance ($\mathit{igd}$) of the four experiments carried out.

\begin{table}
\centering
\caption{Results of 50 evolutionary replicate runs per each problem. Shown is the average inverted generational distance ($\mathit{igd}$) and standard deviation}\label{tab:results}
\begin{tabular}{|r|c|}
\hline
Problem & Inverted generational distance \\ \hline
ZDT1 & 2.06E-04 (2.12E-05)  \\ 
ZDT2 & 2.65E-04 (3.82E-05) \\ 
ZDT3 & 3.81E-02 (2.48E-02) \\
ZDT4 & 1.23E-03 (4.53E-04) \\ \hline
\end{tabular}
\end{table}

We compared our results with two very recent studies. 
Cheng et al. \cite{cheng2017novel} presented HTL-MOPSO, 
a novel hybrid teaching learning based particle swarm optimization algorithm (HTL-PSO) with circular crowded sorting (CCS).
They compared their algorithm to: 
NSGA-II \cite{deb2002fast}, 
TV-MOPSO \cite{tripathi2007multi},
MOEA/D \cite{zhang2007moea},
and MOTLBO \cite{zou2013multi}.

Han et al. \cite{han2018improved} presented an improved multiobjective quantum-behaved particle swarm optimization algorithm based on double search strategy and circular transposon mechanism (MOQPSO-DSCT). They compared their algorithm to:
MPSO/D \cite{dai2015new},
MOQPSO-AG \cite{shi2011multi},
dMOPSO \cite{zapotecas2011multi},
SMPSO \cite{nebro2009smpso},
NSGA-II \cite{deb2002fast},
and MOEA/D\cite{zhang2007moea}.

\begin{figure*}
\centering
\begin{tabular}{cc}
\includegraphics[height=0.45\textwidth]{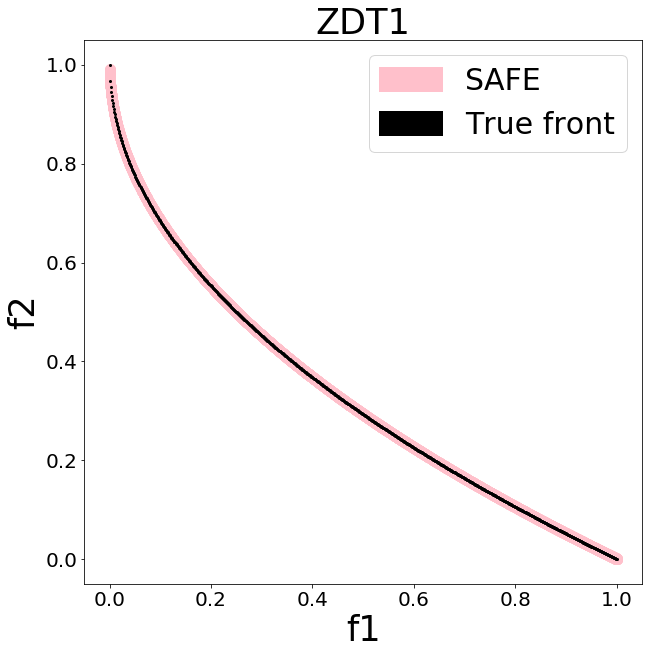} &
\includegraphics[height=0.45\textwidth]{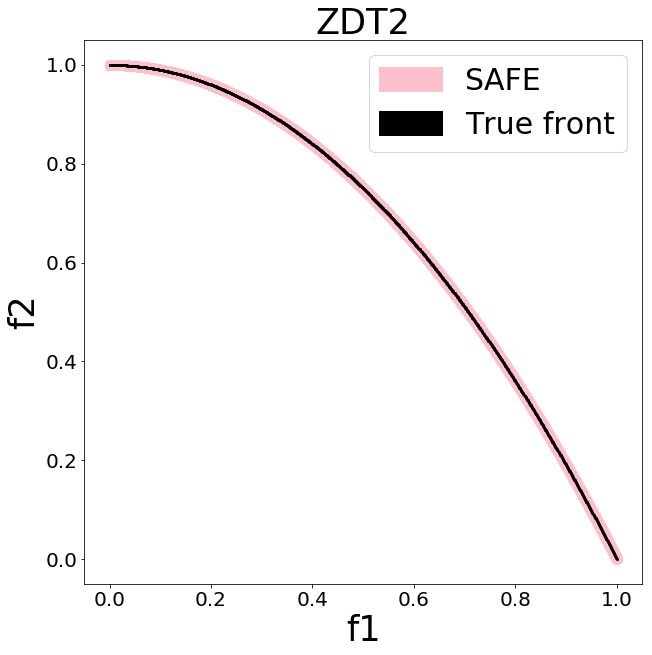} \\
\includegraphics[height=0.45\textwidth]{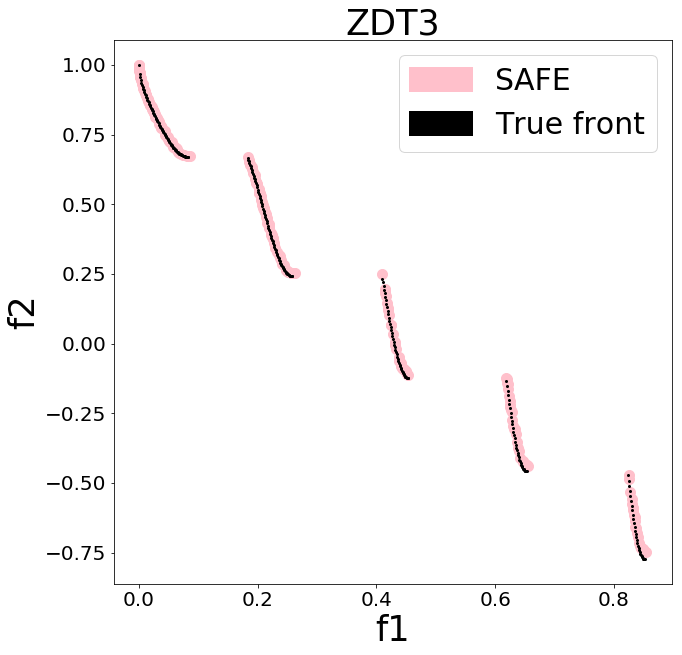} &
\includegraphics[height=0.45\textwidth]{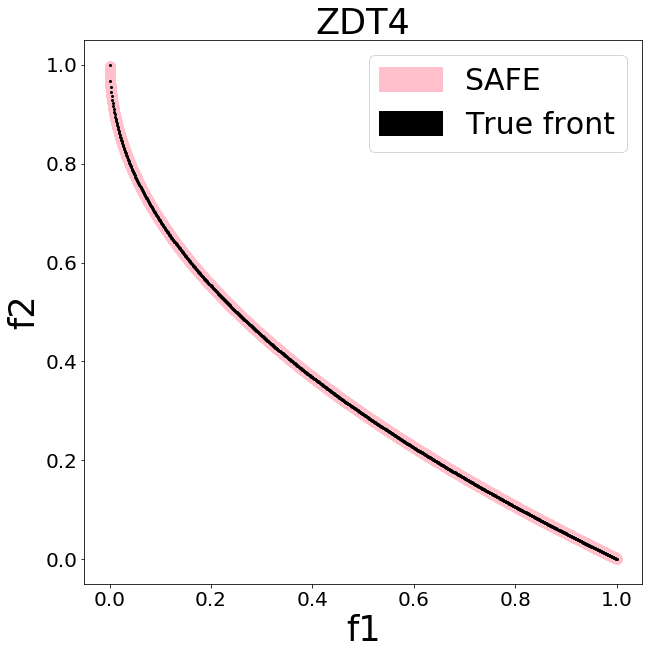} \\
\end{tabular}
\caption{Four sample solutions produced by SAFE, compared with the true fronts. $\mathit{igd}$ values are: 
ZDT1: 1.67E-04,
ZDT2: 1.60E-04,
ZDT3: 5.24E-03,
ZDT4: 6.21E-04.
} \label{fig:fronts}
\end{figure*}

Table~\ref{tab:compare} compares our average $\mathit{igd}$ results with the best values reported in these two recent works, showing the merits of SAFE.
Figure~\ref{fig:fronts} shows four Pareto fronts produced by SAFE, compared with the optimal ground truth Pareto fronts as defined by the ZDT functions. 

\begin{table}
\centering
\caption{Comparing SAFE's average $\mathit{igd}$ results with the best values reported in \cite{cheng2017novel} and \cite{han2018improved}}
\label{tab:compare}
\begin{tabular}{|r|c|c|c|}
\hline
Problem & SAFE & Cheng et al. \cite{cheng2017novel} & Han et al. \cite{han2018improved} \\ \hline
ZDT1 & \textbf{2.06E-04} & 3.88E-03 & 2.81E-03 \\ 
ZDT2 & \textbf{2.65E-04} & 3.85E-03 & 3.92E-03 \\ 
ZDT3 & 3.81E-02 & 4.82E-03 & \textbf{4.45E-03} \\
ZDT4 & \textbf{1.23E-03} & 3.99E-03 & 3.77E-03 \\ \hline
\end{tabular}
\end{table}

We find that by allowing objective function weights to evolve in the SAFE framework, driven by `genotypic novelty', we are able to identify a Pareto optimal solution set that is quite close to the ground-truth optimal solutions across the full span of the front. This behavior makes sense since SAFE evolutionary pressures for objective functions are driven by genotypic novelty, which translates in the ZDT problems to different objective weightings. This diverse set of objective weightings in turn is likely to encourage the evolution of optimal solutions at different locations along the Pareto optimal front. 

\section{Concluding Remarks}
\label{sec:conc}

Aiming to confront the optimization conflation problem---where the objective is conflated with the objective function---we separated these two entities into two populations, and presented SAFE, a coevolutionary algorithm to evolve the two simultaneously. Our previous work showed promising results in a maze domain \cite{Sipper2019} and herein we presented preliminary results that show promise in a multiobjective-optimization domain while simultaneously demonstrating that the SAFE concept of coevolving solutions with objective functions is able to identify a range of Pareto optimal, non-dominated solutions without adopting a Pareto front for fitness evaluation or parent selection. This finding interestingly suggests that SAFE can not only adapt to deceptive fitness landscapes but to problems that demand multiple objectives be optimized. It also presents an alternative strategy to generating non-dominated multiobjective solutions outside of the Pareto front. 

We realize that this work is preliminary and, hoping it leads to new frontiers, have several avenues we wish to explore.

\begin{itemize}
\item The coevolutionary dynamics engendered by SAFE in solving multiobjective problems are likely to be worthy of study in and of themselves.

\item We will adapt the SAFE concept to new problem domains, particularly those that target the machine learning modeling of complex pattens of association in biomedical data mining.  In general, we plan to identify those problems wherein the SAFE concept is not only competitive but clearly advantageous. It is of particular interest to demonstrate whether SAFE can adapt itself to problems where little prior knowledge exists regarding either the objective or the best path to said objective. 

\item We did not discuss computational cost herein, given the preliminary investigative nature of this work, and our interest in simply showing that SAFE is a viable algorithm that produces satisfactory results. Of course, computational cost needs to be addressed in the future. Note that comparing multiobjective algorithms is often far from trivial, since many exhibit various ``hidden'' costs, e.g., hypervolume-based algorithms compute a costly hypervolume measure \cite{antonio2018coevolutionary} (e.g., Bezerra et al. \cite{Bezerra2018} recently noted that, ``The drawback of indicator-based approaches is the computational complexity of some quality indicators such as the hypervolume, which is exponential in the number of objectives in the worst case'').

\item We used a simple measure to drive objective-function evolution in this problem domain---genotypic novelty. Other, possibly better measures might be designed. 

\item Examine the incorporation of alternative, more sophisticated evolutionary algorithm components into the SAFE concept (e.g., selection, elitism, genetic operators, and solution representation).  
\end{itemize}

\bibliographystyle{IEEEtran}
\bibliography{IEEEabrv,bibfile}

\end{document}